\documentclass[10pt]{article}
\usepackage[margin=0.8in]{geometry}
\usepackage{times}
\usepackage{graphicx,booktabs,tabularx,array,float}
\usepackage[breaklinks=true,colorlinks=true,urlcolor=blue]{hyperref}
\usepackage{microtype}
\title{When Reconstruction Helps Event-Camera Person Triggers}
\author{Yifan Shuai\\LiAuto\\\texttt{sddsyf@gmail.com}}
\date{November 10, 2026}
\begin{document}
\maketitle

\begin{abstract}
Event cameras record asynchronous changes in brightness. To use a general visual model as a recording trigger, a system must represent these measurements in a form the model can interpret. We compare direct grayscale accumulation, two temporal encodings, and two intensity reconstructors on paired real event data. On 192 windows from four held-out DSEC recordings, thresholds calibrated only on development data yielded 40 of 52 person detections and 12 of 140 background positives with E2VID. The strongest direct temporal encoding at an admissible fixed cutoff yielded 14 detections and seven background positives. A retrospective comparison at a common 10\% background-positive cap yielded 40 detections for E2VID and at most 19 for any direct encoding. FireNet+ had lower preparation cost and produced 21 detections with zero observed background positives at its frozen cutoff. On the same 192 windows, Qwen3-VL-4B yielded 33/52, 2/140 with E2VID and 9/52, 9/140 with the highest-recall direct rendering, the three-bin image (person detections, background positives). A 96-second exploratory replay illustrates the sensitivity of episode capture and recording starts to cutoff selection. The experiments identify a detection benefit from reconstruction under controlled input history and distinguish it from the cost and false starts of continuous triggering.

\end{abstract}
\section*{1. Introduction}
A visual recording trigger needs to decide when to start a more expensive capture path. An event camera can supply the trigger with a stream of timestamped brightness changes. Those changes can be rendered as a polarity image or reconstructed into an approximation of scene intensity. A frame-trained model may make different decisions on the two inputs even when both originate from the same events.

Existing studies established intensity reconstruction as an interface between event sensors and frame-based vision [1--4]. Here we study its effect on a recording-trigger decision. We ask whether reconstruction improves person presence at a low background-positive rate, whether gains survive a change of recording, and what happens once decisions are evaluated as starts on a timeline. Person presence gives a concrete, labelable task. It is a proxy for one component of a recording policy. An assessment of relevance to a specific wearer requires a separate target and annotation protocol.

An earlier study examined encoder readout and operating-point sensitivity on DailyDVS action clips [18]. The present study evaluates reconstruction and direct representations for person presence on DSEC, using a different task and evaluation set.

The empirical contributions are an equal-history comparison on genuine event data, a frozen-threshold temporal-encoding comparison on four unused recordings, and a continuous replay that separates window classification, episode capture, background starts, and processing time. A small handheld-camera transfer check provides secondary evidence for a different sensor.

\section*{2. Related Work}
\textbf{Event reconstruction and downstream recognition.} E2VID demonstrated that recurrent intensity reconstruction can make event streams usable by existing frame-based classifiers [1]. FireNet reduced reconstruction cost through a smaller recurrent architecture [2]. Cho et al. jointly studied intensity reconstruction and event-based recognition [3]. EZSR distinguishes reconstruction-based transfer to CLIP from approaches that learn an event encoder [4]. EventGPT and EventDrive develop systems trained to incorporate event data into multimodal reasoning [5, 6]. We use fixed visual decision models and change the input representation at evaluation time.

\textbf{Person detection and wearable sensing.} DSEC and DSEC-Detection provide driving events, synchronized imagery and tracks [7, 8]. PEDRo supplies people annotations on a moving handheld DAVIS sensor [9]. OpenGlass evaluates on-device gesture recognition [10]. *Reading in the Dark* reconstructs foveated first-person event inputs for multimodal text recognition [11]. Our main recordings are from vehicle-mounted sensors. The wearable setting motivates reporting background starts and processing costs.

\section*{3. Input and Evaluation}
\subsection*{3.1 Causal event representations}
For each decision time, all event inputs end at that time. A polarity image accumulates 30 ms of events in two channels, mapping counts to red and blue with a fixed saturating transform. A direct grayscale representation merges polarities and darkens pixels as event count increases. The grayscale and polarity maps both retain the full sensor field of view. Two direct temporal controls use the same causal events. The polarity-separated time surface gives each pixel an exponentially decaying value based on its most recent event, with time constant one third of the input history [15]. The three-bin encoding maps early, middle and late event counts to the red, green and blue channels [16]. Each condition supplies one full-field image to the model, with a legend for the representation. Each 640 by 480 image is padded to 640 by 640. Thus these are comparisons of representation-and-legend combinations; prompt wording is a remaining possible contributor.

Reconstruction uses the published E2VID Lightweight checkpoint [1]. Each 30-ms slice is converted to a signed five-bin event tensor using temporal interpolation and nonzero-value normalization. For window evaluation, we initialize the recurrent state and update it with up to ten contiguous slices. The equal-history grayscale control uses exactly the same available event interval, including shorter intervals at recording startup. For replay, state persists within each segment. A lower-resolution variant pools coordinates by a factor of two before voxelization and resizes the reconstructed image to the same visual-model input size. Both variants share weights and parameter count. For the independent window comparison, FireNet+ uses the authors' published \texttt{firenet\_all\_cts} weights [2, 17]. It receives the same 30-ms slices as E2VID, with its upstream unnormalized signed voxel convention. The output is clipped to the intensity range before conversion to an image.

Jev 1.14.0 returns a structured score for whether the image contains an identifiable person or rider [12]. We use the same question across inputs and change the representation description accordingly. Qwen3-VL-4B and Qwen3.5-9B use local 4-bit weights and the relative likelihood of the single answer tokens A (yes) and B (no) [13, 14]. A supervised 38,673-parameter convolutional heatmap model uses event counts directly. It was trained on 1,232 windows from eight DSEC recordings outside the calibration and replay groups. This baseline represents one compact model and training setup.

\subsection*{3.2 Paired windows and recording transfer}
We calibrated thresholds on 64 windows (16 positive and 48 background) from five development recordings. The 10\% constraint permits at most four development background positives; it is an empirical selection rule and does not guarantee a 10\% population false-positive rate. For each condition, the threshold maximized detected person windows subject to at most 10\% development background positives. Ties favored fewer background positives and then the higher threshold. Decision models and reconstruction weights remained fixed.

The paired DSEC set has 48 positive and 128 background windows from four other recordings. Official person and rider tracks were checked against synchronized photos before scoring. Each retained positive window corresponds to one selected track. The synchronized photographs also provide a separate input control. They were excluded from the event-only inputs. Three previously selected startup windows lacked a complete 90-ms history and were omitted before the paired comparison.

The reconstruction comparison was added after initial polarity and grayscale accumulation trials. For the recording-transfer experiment, we froze the 30-ms polarity-map and up-to-300-ms reconstruction Jev cutoffs at 0.63 and 0.43 and transferred them to four recordings outside the calibration and paired groups. Those recordings provided 36 positive and 81 background windows. Some were used in earlier project studies. The recordings are separate from this study's cutoff calibration.

\subsection*{3.3 Frozen temporal controls on unused recordings}
We compared the two direct temporal encodings at 90 ms and up to 300 ms on the earlier paired set. We then evaluated a primary 300-ms comparison on four DSEC recordings outside all earlier calibration and comparison groups. They contribute 52 person tracks and 140 background windows. From 234 annotation-based candidates we retained 192 after score-blind review of the paired reference photos by an AI assistant. No independent human annotation study or inter-rater agreement measurement was performed. The 42 exclusions include visibility, ambiguity and duplicate-target checks, so the resulting sample estimates performance conditional on this selection process. Positive examples contribute one window per retained track. Background anchors are at least 350 ms apart and at least 150 ms from official person annotations. All retained histories are exactly 300 ms. The images contain no reference boxes. Thresholds for the new input conditions came from the same 64 development windows and were fixed before fresh scoring.

\subsection*{3.4 Continuous replay and handheld transfer}
We replayed four unbroken 24-second segments from the paired DSEC recording group. Segment selection used only track annotations and favored intervals containing both positive and background decisions. Two separate 24-second development segments provided cutoffs under the same 10\% decision-level constraint. At each 480-ms decision tick, a nearby annotation gave person, background, or unknown status. Person and rider boxes had to be at least 8 by 24 pixels after clipping. Frames containing only smaller person boxes and frames without a nearby annotation were unknown. References and the decision grid were fixed before scoring.

An episode is a run of positive ticks. A recording start occurs at an upward threshold crossing, subject to a two-second cooldown. A start within a positive run captures that episode. Background and unknown starts are counted separately. Grid delay runs from the first positive tick to the start and excludes inference and communication. We report start counts alongside the replay duration.

PEDRo provides a secondary positive-only transfer check on a handheld DAVIS346. We selected the middle published sample from 12 separate official test recordings. Polarity accumulation used the final released sample, and reconstruction used up to eight consecutive indexed samples. Actual durations varied. State was reset across timestamp gaps over 5 ms. Both Qwen3-VL-4B thresholds came from DSEC. Every selected PEDRo sample contained a labeled person, so this check estimates neither false starts nor continuous trigger performance.

\section*{4. Results}
\subsection*{4.1 Reconstruction changes the decision at fixed event history}
\begin{table}[htbp]\centering\footnotesize\setlength{\tabcolsep}{3pt}
\begin{tabularx}{\linewidth}{>{\raggedright\arraybackslash}X>{\raggedright\arraybackslash}X>{\raggedright\arraybackslash}X>{\raggedright\arraybackslash}X>{\raggedright\arraybackslash}X}
\toprule
Input & Decision model & Cutoff & Person detections & Background positives \\
\midrule
30-ms polarity map & Jev & 0.63 & 11/48 & 0/128 \\
30-ms polarity map & Qwen3-VL-4B & 0.00070967 & 10/48 & 1/128 \\
30-ms polarity map & Qwen3.5-9B & 0.05340333 & 12/48 & 1/128 \\
90-ms grayscale accumulation & Jev & 0.53 & 10/48 & 3/128 \\
90-ms reconstruction & Jev & 0.59 & 29/48 & 2/128 \\
Grayscale accumulation, up to 300 ms & Jev & 0.23 & 15/48 & 18/128 \\
E2VID, up to 300 ms & Jev & 0.43 & \textbf{35/48} & \textbf{2/128} \\
Grayscale accumulation, up to 300 ms & Qwen3-VL-4B & 4.540e-5 & 11/48 & 12/128 \\
Time surface, up to 300 ms & Qwen3-VL-4B & 1.01 & 0/48 & 0/128 \\
Three-bin image, up to 300 ms & Qwen3-VL-4B & 2.754e-5 & 5/48 & 19/128 \\
E2VID, up to 300 ms & Qwen3-VL-4B & 0.00192673 & \textbf{28/48} & \textbf{0/128} \\
Synchronized photo & Jev & 0.61 & 36/48 & 0/128 \\
Synchronized photo & Qwen3.5-9B & 0.13296422 & 42/48 & 0/128 \\
\bottomrule\end{tabularx}
\par\smallskip Table 1. Paired DSEC window results. Each input-model pair uses its own development-only cutoff, displayed rounded. The Qwen time-surface cutoff exceeded the score range because the development rule found no positive detection within budget.
\end{table}
At equal event history, Jev gained 20 person detections with 16 fewer background positives. On the same up-to-300-ms histories, Qwen3-VL-4B detected 28/48 with E2VID versus 11/48 with direct grayscale accumulation. The paired comparison with synchronized photographs gives 31 detections by both Jev inputs, five with photos alone, four with reconstruction alone and eight missed by both. The similar totals therefore conceal distinct errors.

On the four recording-transfer groups, polarity images detected 10/36 and reconstruction detected 27/36. Both yielded 2/81 background positives. The recall gain was visible in every transfer recording, including a nighttime sequence with 1/10 detections on polarity images and 7/10 after reconstruction. The 12-case handheld transfer found 10 persons with polarity images and 12 with reconstructed images. This last result is descriptive because there is one sampled positive per recording and no background control.

\begin{figure}[htbp]\centering\includegraphics[width=\linewidth,height=0.52\textheight,keepaspectratio]{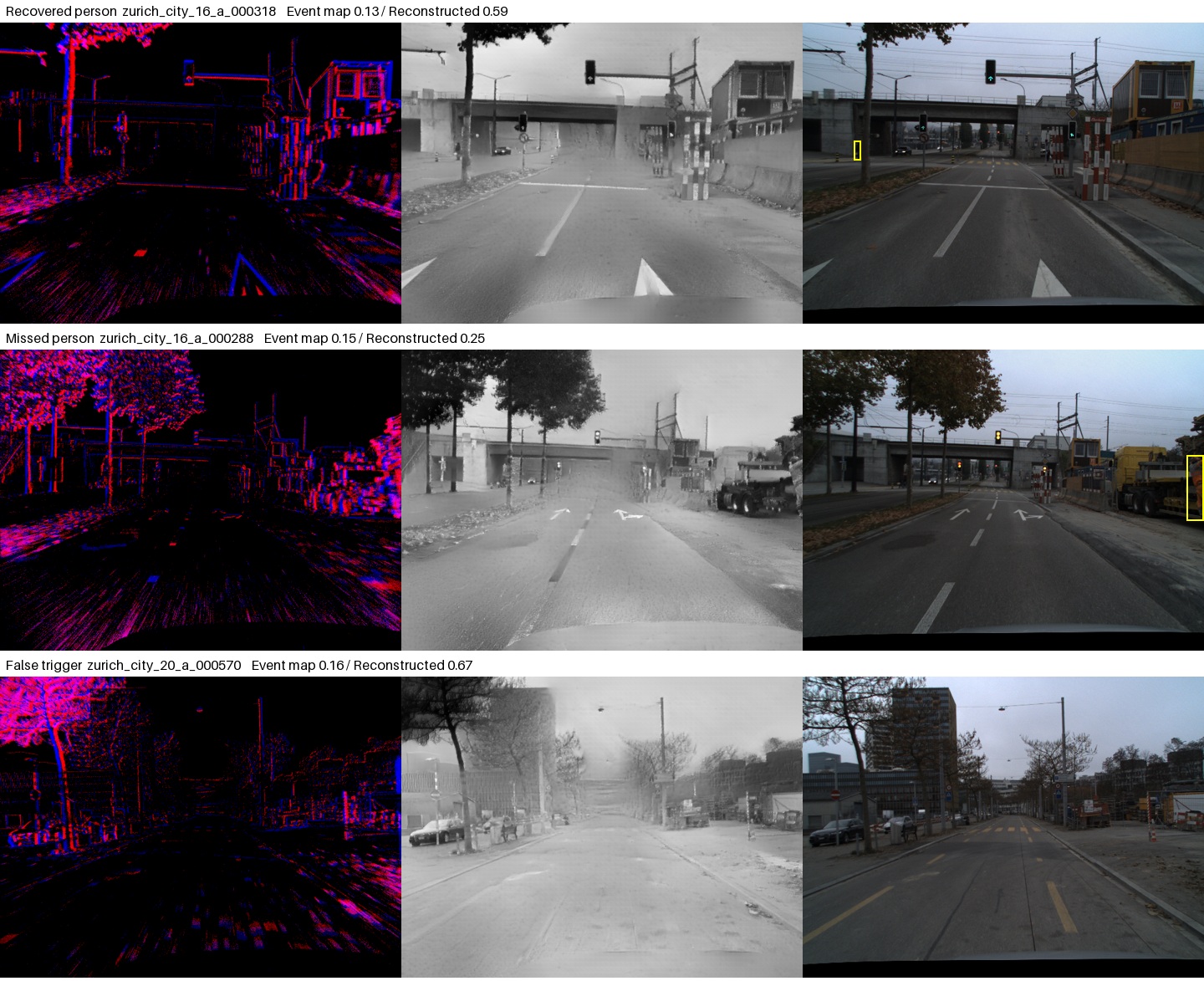}
\par\smallskip\footnotesize Figure 1. One recovered person-positive window, one missed window and one background positive. Columns show the raw polarity rendering, the E2VID output, and a synchronized photograph for reference. The yellow box marks the sampled reference target. A positive response scores person presence anywhere in the frame.
\end{figure}
\subsection*{4.2 Reconstruction outperforms the tested direct temporal encodings}
\begin{table}[htbp]\centering\footnotesize\setlength{\tabcolsep}{3pt}
\begin{tabularx}{\linewidth}{>{\raggedright\arraybackslash}X>{\raggedright\arraybackslash}X>{\raggedright\arraybackslash}X>{\raggedright\arraybackslash}X>{\raggedright\arraybackslash}X}
\toprule
300-ms representation with Jev & Frozen cutoff & Person detections & Background positives & Retrospective detections and background positives at the 10\% cap \\
\midrule
Grayscale accumulation & 0.23 & 18/52 & 31/140 & 11/52, 13/140 \\
Polarity-separated time surface & 0.29 & 11/52 & 11/140 & 12/52, 12/140 \\
Three-bin temporal image & 0.15 & 14/52 & 7/140 & 19/52, 14/140 \\
FireNet+ & 0.69 & 21/52 & 0/140 & 36/52, 14/140 \\
E2VID & 0.43 & \textbf{40/52} & \textbf{12/140} & \textbf{40/52, 12/140} \\
\bottomrule\end{tabularx}
\par\smallskip Table 2. Identical 300-ms histories on four previously unused DSEC recordings. The last column scans test scores for diagnostic purposes. Frozen cutoffs in the first three result columns were calibrated before the test.
\end{table}
At their frozen cutoffs, E2VID recovered 28 positives missed by the three-bin image and missed two that it captured. E2VID detected more person windows than the direct temporal encodings in three recordings and tied them at 7/10 in the night recording. The pooled recall difference was 50.0 percentage points, with a leave-one-recording-out range of 40.5 to 61.9 points. A descriptive paired bootstrap over the four recordings gave 95\% percentile ranges of 18.2 to 69.0 points for recall and -1.4 to 8.7 points for background-positive rate. Each bootstrap draw resampled four whole recordings with replacement (256 equally weighted ordered draws). With only four groups, these ranges describe within-sample sensitivity. At a retrospective cap of seven background positives, E2VID captured 37/52 with four background positives. The three-bin image captured 14/52 with seven. These diagnostic points do not change the frozen cutoff results.

FireNet+ used 37,777 parameters compared with E2VID's 10.71 million. In the matched host benchmark, FireNet+ prepared an image in 0.48 seconds versus 2.03 seconds for E2VID. Its frozen Jev cutoff captured 21/52, with 0/140 observed background positives. The retrospective scan under the common 10\% cap yielded 36/52 for FireNet+ and 40/52 for E2VID. The four recordings cannot establish that the two methods are equivalent. The timings describe host image preparation.

On the earlier 90-ms comparison, Jev found 12/48 from time surfaces, 13/48 from three-bin images, and 29/48 from E2VID. On the earlier up-to-300-ms comparison, the same inputs yielded 4/48, 16/48, and 35/48 respectively. On the same up-to-300-ms inputs, Qwen3-VL-4B found 28/48 with E2VID at zero observed background positives. The best direct rendering at an admissible frozen cutoff was grayscale accumulation with 11/48 and 12/128 background positives.

\begin{figure}[htbp]\centering\includegraphics[width=\linewidth,height=0.52\textheight,keepaspectratio]{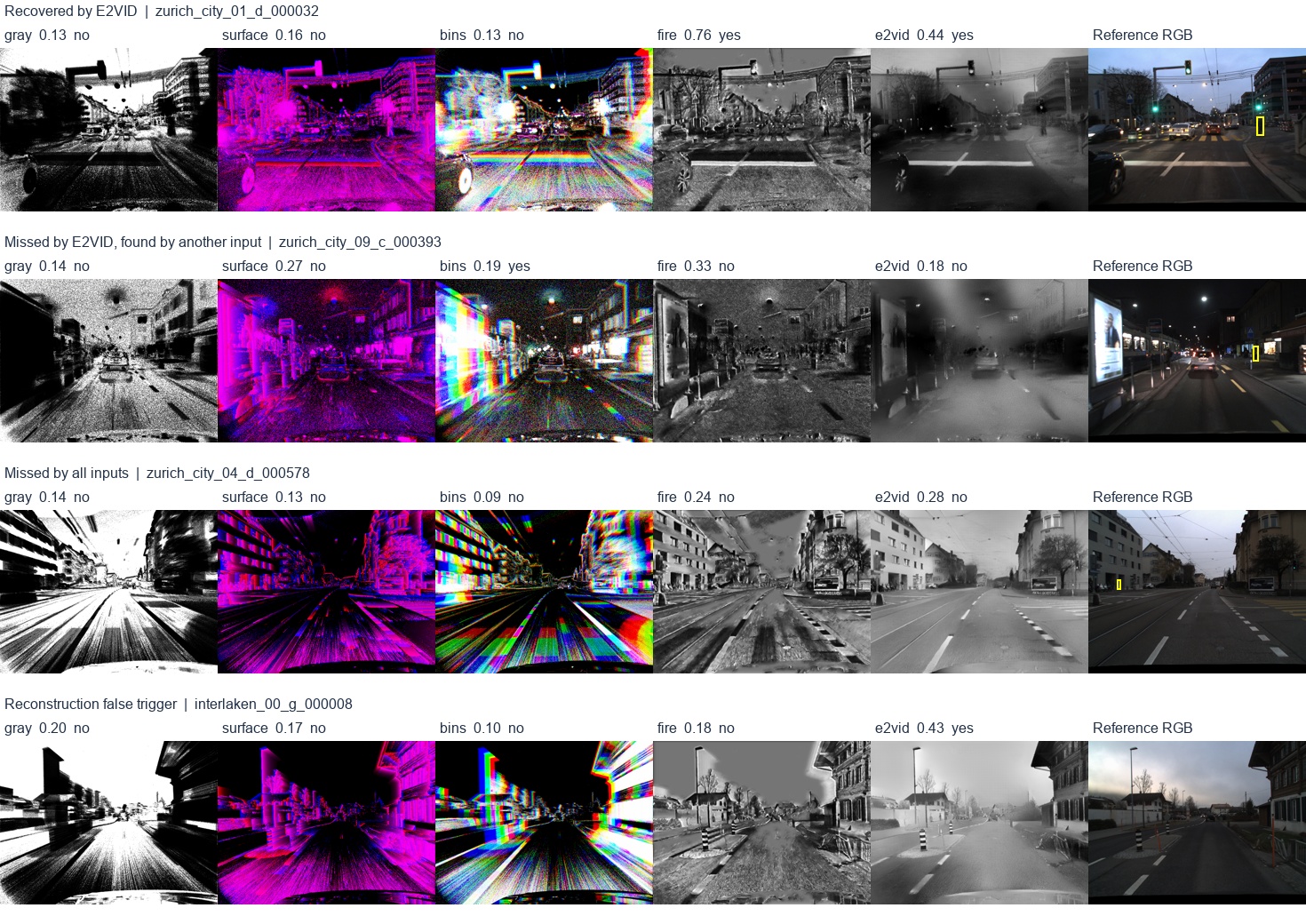}
\par\smallskip\footnotesize Figure 2. Input representations derived from identical 300-ms intervals. The rightmost photograph serves as a reference only. Scores and decisions use the respective frozen cutoffs. Examples illustrate both recovered and missed positives and a background trigger.
\end{figure}
\subsection*{4.3 Open-weight replication and screening sensitivity}
We scored the same 192 retained windows with the pinned Qwen3-VL-4B checkpoint and five matched 300-ms images. Four cutoffs were reused from the previously frozen development-only policy. The FireNet+ cutoff used the same 64 development windows and selection rule and was fixed before any of the 192 replication windows were scored.

\begin{table}[htbp]\centering\footnotesize\setlength{\tabcolsep}{3pt}
\begin{tabularx}{\linewidth}{>{\raggedright\arraybackslash}X>{\raggedright\arraybackslash}X>{\raggedright\arraybackslash}X>{\raggedright\arraybackslash}X>{\raggedright\arraybackslash}X}
\toprule
Input & Frozen Qwen cutoff & Retained 192 & All 234, original labels & Retained, retrospective 10\% cap \\
\midrule
Grayscale accumulation & 4.5397861e-05 & 5/52, 19/140 & 7/72, 27/162 & 5/52, 9/140 \\
Polarity-separated time surface & 1.01 & 0/52, 0/140 & 0/72, 0/162 & 11/52, 14/140 \\
Three-bin temporal image & 2.75357e-05 & 9/52, 9/140 & 11/72, 12/162 & 9/52, 5/140 \\
FireNet+ & 0.010986944 & 18/52, 1/140 & 24/72, 2/162 & 34/52, 10/140 \\
E2VID & 0.0019267347 & 33/52, 2/140 & 43/72, 2/162 & 41/52, 8/140 \\
\bottomrule\end{tabularx}
\par\smallskip Table 3. Counts show person detections followed by background positives. The fourth column includes all 42 previously screened-out candidates using original dataset-derived references at the same frozen cutoff. The last column is a diagnostic scan of the 192 retained test scores.
\end{table}
On the retained sample, E2VID had higher recall than the three-bin image on three recordings and tied on the night recording. FireNet+ yielded 18/52, 1/140 at its frozen cutoff. The retrospective 10\% cap gave 41/52, 8/140 for E2VID, 34/52, 10/140 for FireNet+, and at most 11/52 detections for the direct renderings. Qwen3-VL-4B with E2VID fell short of the 70\% recall target at its frozen cutoff.

The annotation-only sensitivity check added all excluded candidates and changed neither the rendering procedure nor the thresholds. This comparison tests dependence on assistant screening under the original annotation rule. It cannot resolve uncertain human presence, missed annotations or possible duplicate encounters. The 192-window analysis remains the screened primary set. For E2VID and grayscale accumulation, the corresponding all-candidate counts were 43/72, 2/162 and 7/72, 27/162.

The existing supervised 30-ms compact CNN scored 3/52, 1/140 on the same retained windows at its original development-only cutoff (0.94751787). Its input duration and supervised training distinguish it from the matched 300-ms representation comparison. It serves as a deployment-oriented reference.

\subsection*{4.4 Continuous decisions reveal a different operating trade-off}
\begin{table}[htbp]\centering\footnotesize\setlength{\tabcolsep}{3pt}
\begin{tabularx}{\linewidth}{>{\raggedright\arraybackslash}X>{\raggedright\arraybackslash}X>{\raggedright\arraybackslash}X>{\raggedright\arraybackslash}X>{\raggedright\arraybackslash}X>{\raggedright\arraybackslash}X>{\raggedright\arraybackslash}X}
\toprule
Input and decision model & Person ticks detected & Background ticks positive & Episodes captured & Background starts & Unknown starts & Median grid delay \\
\midrule
30-ms polarity + Jev & 42/88 & 26/96 & 8/9 & 12 & 1 & 1.44 s \\
300-ms grayscale accumulation + Jev & 0/88 & 2/96 & 0/9 & 2 & 0 & --- \\
640$\times$480 recurrent reconstruction + Jev & 60/88 & 20/96 & 7/9 & 7 & 1 & 0 s \\
320$\times$240 recurrent reconstruction + Jev & 42/88 & 6/96 & 8/9 & 3 & 2 & 0.96 s \\
Compact event CNN & 23/88 & 8/96 & 6/9 & 4 & 1 & 0.24 s \\
\bottomrule\end{tabularx}
\par\smallskip Table 4. Continuous test replay across 96 seconds and 200 ticks. Sixteen ticks had unknown references. The episode delays are medians over captured episodes only. Zero means a start at the first positive tick.
\end{table}
With cutoffs independently calibrated by the same replay-development rule, the polarity map and half-resolution reconstruction each captured 8/9 episodes, with 12 and three background starts. Their achieved background-tick positive rates differed (26/96 versus 6/96). The full-resolution reconstruction had 20/96 background-positive ticks. Applying the earlier window cutoffs instead, the polarity map and full-resolution reconstruction each produced one background start and captured 6/9 and 7/9 episodes. The selected 96 seconds and nine episodes support an exploratory comparison. Cooldown makes successive starts dependent, and a per-hour extrapolation would obscure the short exposure. The compact CNN captured 6/9 episodes with four background starts under the replay cutoff. These counts are too small to establish a reliable advantage over the CNN. FireNet+ was tested on fixed windows and was not included in this replay.

\begin{figure}[htbp]\centering\includegraphics[width=\linewidth,height=0.52\textheight,keepaspectratio]{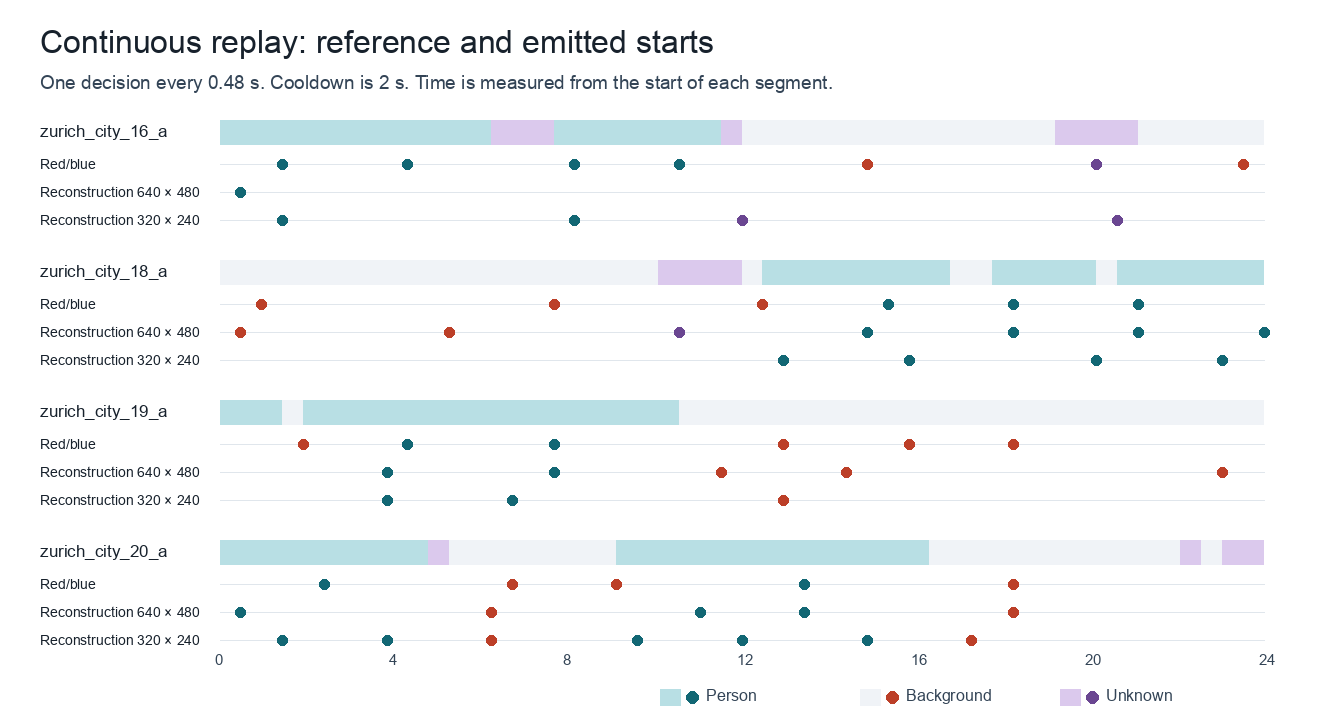}
\par\smallskip\footnotesize Figure 3. Trigger starts from three Jev inputs at the continuous-development cutoffs. Color encodes the reference at each start. Starts with unknown reference are retained as a separate category.
\end{figure}
\subsection*{4.5 Host time and throughput}
On an Apple M1 Pro, the median voxelization, model and synchronization time for each 30-ms recurrent update was 112.3 ms at native resolution and 37.2 ms at half resolution. Accumulated processing took 3.83 and 1.29 seconds per second of source input, respectively. These measurements exclude the service call. The median Jev call for the half-resolution variant was 1.03 seconds. The compact CNN's median CPU step was 2.8 ms after event counts were available. The continuous replay uses an offline decision clock, and the measured implementations operate slower than that clock for the full reconstruction-and-remote-decision path.

In a synchronized preparation-only benchmark on 16 matched in-memory development cases, E2VID and FireNet+ took 2.03 and 0.48 seconds respectively to process the ten 30-ms steps and retrieve the final intensity array on the same M1 Pro. The models contain 10.71 million and 37,777 parameters. These times characterize the tested host implementation and omit decision-model latency and device energy.

\section*{5. Discussion}
The equal-history result shows that longer event accumulation alone does not explain the observed detection gain. The advantage over the two temporal renderings persists at the same retrospective false-positive cap. The comparison concerns the tested renderings and fixed visual models. Learned event-native alternatives and other temporal encodings may behave differently. A plausible interpretation is that intensity reconstruction converts sparse motion-dependent edges into a representation better aligned with frame-trained models. We did not isolate the effects of temporal integration, the pretrained reconstruction prior, or output contrast.

For an eyewear design, the useful trade-off is captured episodes versus unwanted starts and processing cost. A configuration with good isolated-window recall can behave differently once repeated scores, persistent positives, and cooldown enter the policy. The half-resolution configuration reduced reconstruction processing time. Its measured throughput still falls short of the event-update cadence, and reducing resolution preserves the number of model weights.

The four continuous segments are short, were selected for annotation coverage, and reuse recordings from the paired experiment. Official driving tracks may miss people, particularly outside the apparent annotation scope. Thus the replay estimates annotation-relative background starts. The separate recording transfer uses window samples. The small handheld check covers positive windows only. The independent window set still shares the DSEC sensor family and includes nearby routes. All evaluations concern visible people, while real capture decisions may also depend on attention, sound or the type of event. Glasses-mounted recordings with independently annotated onsets, longer negative runs and synchronized hardware power measurements are needed for an always-on claim.

The small development set makes cutoff estimates unstable. The four fresh recordings are the units of environmental replication; nearby windows and tracks cannot be treated as independent environments. We report recording-specific counts and leave-one-recording-out direction checks instead of claiming population significance from 192 independent trials. The retrospective test-score scans are optimistic diagnostics. The fresh comparisons cover one commercial service and one open-weight model. The 234-window sensitivity analysis uses inherited annotations for ambiguous or duplicate cases and is not an independent human-label audit. Service revisions, stochastic response variability and pretraining overlap with public datasets remain uncontrolled. The simple renderings were not exhaustively optimized. Consequently, the results support a scoped empirical comparison, with novelty in the controlled trigger evaluation rather than in the reconstruction algorithm itself.

\section*{6. Conclusion}
In paired real DSEC data, E2VID reconstruction improved the tested visual person decisions at equal event history. The benefit transferred to four other recordings under frozen thresholds and was also observed with a public vision-language model. Continuous replay exposed a cost and false-start trade-off that the window results alone could not predict. The accompanying score tables, decision cutoffs, sample identifiers and calculation script permit independent recomputation of the primary comparison.

\section*{Appendix A. Reproduction details}
The primary Jev question was held fixed as follows.

``Does the attached event-camera image contain a clearly identifiable actual person or human body outline? Count pedestrians and people riding bicycles. Count partially visible people only if clearly recognizable. Exclude isolated ambiguous marks, printed depictions and reflections.''

The API model identifier was \texttt{jev-1.14.0}; the response field \texttt{person\_present.noul} supplied the score. It is used as a ranking score, without a probability-calibration claim. Qwen receives the same question followed by ``Reply with exactly A for yes or B for no. Do not explain.'' Its score is the softmax over the first-token log probabilities of A and B, with thinking disabled. The local models were \texttt{mlx-community/Qwen3-VL-4B-Instruct-4bit} at revision \texttt{2fd8dacbdb8f1e54b8c005f081ec5bf79c56376b} and \texttt{mlx-community/Qwen3.5-9B-4bit} at revision \texttt{8b2b98c00a6b4d291155e4890773ca8f769aee53}.

For a pixel with n events, direct grayscale intensity is floor(255 exp(-n/3)). Each temporal-bin channel uses floor(255(1-exp(-n/3))). The time surface uses floor(255 exp(-3(t\_end-t\_last)/history)), with zero for missing events, positive polarity in red and negative polarity in blue. E2VID and FireNet+ use ten causal 30-ms updates for the fresh 300-ms inputs. The accompanying protocol lists the channel conventions. Threshold comparison uses score greater than or equal to the frozen cutoff.

\begin{table}[htbp]\centering\footnotesize\setlength{\tabcolsep}{3pt}
\begin{tabularx}{\linewidth}{>{\raggedright\arraybackslash}X>{\raggedright\arraybackslash}X>{\raggedright\arraybackslash}X>{\raggedright\arraybackslash}X>{\raggedright\arraybackslash}X>{\raggedright\arraybackslash}X>{\raggedright\arraybackslash}X}
\toprule
Recording & Person / background windows & Grayscale TP / FP & Time surface TP / FP & Three-bin TP / FP & FireNet+ TP / FP & E2VID TP / FP \\
\midrule
interlaken\_00\_g & 14 / 35 & 4 / 6 & 1 / 1 & 4 / 3 & 7 / 0 & 12 / 4 \\
zurich\_city\_01\_d & 13 / 38 & 4 / 11 & 3 / 5 & 1 / 0 & 5 / 0 & 8 / 4 \\
zurich\_city\_04\_d & 15 / 35 & 4 / 8 & 0 / 0 & 2 / 4 & 6 / 0 & 13 / 3 \\
zurich\_city\_09\_c & 10 / 32 & 6 / 6 & 7 / 5 & 7 / 0 & 3 / 0 & 7 / 1 \\
\bottomrule\end{tabularx}
\par\smallskip Table 5. Recording-level primary test counts. TP denotes a positive reference with a positive decision; FP denotes a background reference with a positive decision.
\end{table}
\section*{Artifact availability and assistance disclosure}
The ancillary archive includes the 960 Jev primary scores, the 192 sample identities and event intervals, 1,170 Qwen scores with screening flags, the compact CNN scores, frozen cutoffs, and a standalone metric-recomputation script. It contains derived measurements. Source event data and pretrained weights are obtained from the cited providers. Full training and inference environments and remote-service reproducibility remain outside this compact artifact. AI assistance was used for experiment implementation, reference-image screening, analysis, manuscript drafting and internal review. This review was part of manuscript preparation and does not constitute independent peer review.

\section*{References}
[1] H. Rebecq, R. Ranftl, V. Koltun, and D. Scaramuzza, ``Events-to-Video: Bringing Modern Computer Vision to Event Cameras,'' CVPR, 2019. \href{https://arxiv.org/abs/1904.08298}{Paper}.

[2] C. Scheerlinck et al., ``Fast Image Reconstruction with an Event Camera,'' WACV, 2020. \href{https://openaccess.thecvf.com/content_WACV_2020/html/Scheerlinck_Fast_Image_Reconstruction_with_an_Event_Camera_WACV_2020_paper.html}{Paper}.

[3] H. Cho et al., ``Label-Free Event-Based Object Recognition via Joint Learning with Image Reconstruction from Events,'' ICCV, 2023. \href{https://openaccess.thecvf.com/content/ICCV2023/papers/Cho_Label-Free_Event-based_Object_Recognition_via_Joint_Learning_with_Image_Reconstruction_ICCV_2023_paper.pdf}{Paper}.

[4] Y. Yang et al., ``EZSR: Event-Based Zero-Shot Recognition,'' CVPR, 2025. \href{https://openaccess.thecvf.com/content/CVPR2025/papers/Yang_EZSR_Event-based_Zero-Shot_Recognition_CVPR_2025_paper.pdf}{Paper}.

[5] S. Liu et al., ``EventGPT: Event Stream Understanding with Multimodal Large Language Models,'' 2024. \href{https://arxiv.org/abs/2412.00832}{Preprint}.

[6] D. Lu et al., ``EventDrive: Event Cameras for Vision-Language Driving Intelligence,'' 2026. \href{https://arxiv.org/abs/2606.18242}{Preprint}.

[7] M. Gehrig, W. Aarents, D. Gehrig, and D. Scaramuzza, ``DSEC: A Stereo Event Camera Dataset for Driving Scenarios,'' IEEE Robotics and Automation Letters, 2021. \href{https://arxiv.org/abs/2103.06011}{Paper}.

[8] D. Gehrig and D. Scaramuzza, ``Low-latency automotive vision with event cameras,'' Nature, vol. 629, pp. 1034--1040, 2024. \href{https://doi.org/10.1038/s41586-024-07409-w}{Paper}.

[9] C. Boretti et al., ``PEDRo: An Event-Based Dataset for Person Detection in Robotics,'' CVPR Workshops, 2023. \href{https://openaccess.thecvf.com/content/CVPR2023W/EventVision/papers/Boretti_PEDRo_An_Event-Based_Dataset_for_Person_Detection_in_Robotics_CVPRW_2023_paper.pdf}{Paper} and \href{https://zenodo.org/records/13331985}{dataset v2.0}.

[10] P. Bonazzi et al., ``OpenGlass: Ultra-Low-Power On-Device AI Eyewear with Event-based Vision,'' 2026. \href{https://arxiv.org/abs/2606.07431}{Preprint}.

[11] C. Brander, G. Cioffi, N. Messikommer, and D. Scaramuzza, ``Reading in the Dark with Foveated Event Vision,'' CVPR Workshops, 2025. \href{https://openaccess.thecvf.com/content/CVPR2025W/EventVision/papers/Brander_Reading_in_the_Dark_with_Foveated_Event_Vision_CVPRW_2025_paper.pdf}{Paper}.

[12] TypeSafe AI, ``Jev: For the builders AND dreamers,'' 2026. \href{https://typesafe.ai/blog/jev-1-14}{Model release}.

[13] Qwen Team, ``Qwen3-VL Technical Report,'' 2025. \href{https://arxiv.org/abs/2511.21631}{Preprint}.

[14] Qwen Team, ``Qwen3.5-9B,'' 2026. \href{https://huggingface.co/Qwen/Qwen3.5-9B}{Model card}.

[15] X. Lagorce et al., ``HOTS: A Hierarchy of Event-Based Time-Surfaces for Pattern Recognition,'' IEEE Transactions on Pattern Analysis and Machine Intelligence, 2017. \href{https://researchportal.hkust.edu.hk/en/publications/hots-a-hierarchy-of-event-based-time-surfaces-for-pattern-recogni/}{Paper record}.

[16] D. Gehrig et al., ``End-to-End Learning of Representations for Asynchronous Event-Based Data,'' ICCV, 2019. \href{https://openaccess.thecvf.com/content_ICCV_2019/html/Gehrig_End-to-End_Learning_of_Representations_for_Asynchronous_Event-Based_Data_ICCV_2019_paper.html}{Paper}.

[17] T. Stoffregen et al., ``Reducing the Sim-to-Real Gap for Event Cameras,'' ECCV, 2020. \href{https://github.com/TimoStoff/event_cnn_minimal}{Author implementation and model weights}.

[18] Y. Shuai, ``Readout and Operating-Point Sensitivity in Event-Stream Candidate Selection,'' 2026. \href{https://arxiv.org/abs/2610.36743}{Preprint}.

\end{document}